\documentclass[journal]{IEEEtran}
\usepackage{amsmath,amsfonts}
\usepackage{algorithmic}
\usepackage{algorithm}
\usepackage{array}
\usepackage[caption=false,font=footnotesize,labelfont=rm,textfont=rm]{subfig}
\usepackage{textcomp}
\usepackage{stfloats}
\usepackage{url}
\usepackage{verbatim}
\usepackage{graphicx}
\usepackage{cite}
\usepackage{booktabs}
\usepackage{bm}                    
\usepackage{amssymb}
\usepackage{hyperref}
\usepackage{orcidlink}
\usepackage{lineno}
\usepackage{tabularray}
\usepackage{multirow}
\usepackage{makecell}
\usepackage{geometry}
\usepackage{ragged2e}
\usepackage{xcolor}
\usepackage{microtype}
\usepackage{siunitx}

\definecolor{col1} {HTML}{FF236F}
\definecolor{col2} {HTML}{3D96F5}
\definecolor{col3} {HTML}{1FBEB5}
\definecolor{col4} {HTML}{241F76}
\definecolor{col5} {HTML}{241F76}
\definecolor{col6} {HTML}{78CB37}
\definecolor{col7} {HTML}{9C9C2B}
\definecolor{col8} {HTML}{CF56A9}
\definecolor{col9} {HTML}{E41AA5}
\definecolor{col10}{HTML}{14DC9F}
\definecolor{col11}{HTML}{39ADEF}
\definecolor{col12}{HTML}{5C44C2}
\definecolor{col13}{HTML}{FDB848}
\definecolor{col14}{HTML}{58986C}
\definecolor{col15}{HTML}{1DDE20}
\definecolor{col16}{HTML}{FE5CF4}
\definecolor{col17}{HTML}{ABB30F}
\definecolor{col18}{HTML}{25AFF0}

\begin{document}


\title{
Data-Augmented Deep Learning for Downhole \linebreak
Depth Sensing and Validation
}

\author{%
    Si-Yu~Xiao      \orcidlink{0009-0006-3095-3242},
    Xin-Di~Zhao     \orcidlink{0009-0007-4304-001X},
    Tian-Hao~Mao    \orcidlink{0009-006-1898-1080},
    Yi-Wei~Wang     \orcidlink{0009-0002-5203-7254},
    Yu-Qiao~Chen    \orcidlink{0009-0000-7241-8850},\linebreak
    Hong-Yun~Zhang  \orcidlink{0009-0007-6462-1075},
    Jian~Wang       \orcidlink{0009-0002-7933-8671},
    Jun-Jie~Wang    \orcidlink{0000-0001-7183-422X},
    Shuang~Liu      \orcidlink{0000-0002-0587-4415},
    Tu-Pei~Chen     \orcidlink{0000-0002-1098-9575} and
    Yang~Liu        \orcidlink{0000-0003-0615-7036}\IEEEauthorrefmark{1}

\thanks{Si-Yu Xiao, Tian-Hao Mao, Yi-Wei Wang, Yu-Qiao Chen, Jun-Jie Wang, Shuang Liu and Yang Liu are with the State Key Laboratory of Electronic Thin Films and Integrated Devices, University of Electronic Science and Technology of China, Chengdu 611731, China.}
\thanks{Xin-Di Zhao, Hong-Yun Zhang and Jian Wang are with the Southwest Branch of China National Petroleum Corporation Logging Co., Ltd., Chongqing 401100, China.}
\thanks{Tu-Pei Chen is with the School of Electrical and Electronic Engineering, Nanyang Technological University, Singapore 639798.}
\thanks{This work is supported by NSFC under project No.~62404033 and 62404034. This work is also supported by China National Petroleum Corporation Logging Co., Ltd. (CNLC) under project No.~CNLC2023-7A01.}
\thanks{\IEEEauthorrefmark{1}Corresponding author.}
}


\maketitle

\begin{abstract}
Accurate downhole depth measurement is essential for oil and gas well operations, directly influencing reservoir contact, production efficiency, and operational safety. Collar correlation using a casing collar locator (CCL) is fundamental for precise depth calibration. While neural network has achieved significant progress in collar recognition, preprocessing methods for such applications remain underdeveloped. Moreover, the limited availability of real well data poses substantial challenges for training neural network models that require extensive datasets. This paper presents a system integrated into a downhole toolstring for CCL log acquisition to facilitate dataset construction. Comprehensive preprocessing methods for data augmentation are proposed, and their effectiveness is evaluated using baseline neural network models. Through systematic experimentation across diverse configurations, the contribution of each augmentation method is analyzed. Results demonstrate that standardization, label distribution smoothing (LDS), and random cropping are fundamental prerequisites for model training, while label smoothing regularization (LSR), time scaling, and multiple sampling significantly enhance model generalization capabilities. Incorporating the proposed augmentation methods into the two baseline models results in maximum F1 score improvements of 0.027 and 0.024 for the TAN and MAN models, respectively. Furthermore, applying these techniques yields F1 score gains of up to 0.045 for the TAN model and 0.057 for the MAN model compared to prior studies. Performance evaluation on real CCL waveforms confirms the effectiveness and practical applicability of our approach. This work addresses the existing gaps in data augmentation methodologies for training casing collar recognition models under CCL data-limited conditions, and provides a technical foundation for the future automation of downhole operations.
\end{abstract}


\newpage

\section{Introduction}

\begin{figure}[!b]
    \centering
    \includegraphics[width=0.75\linewidth]{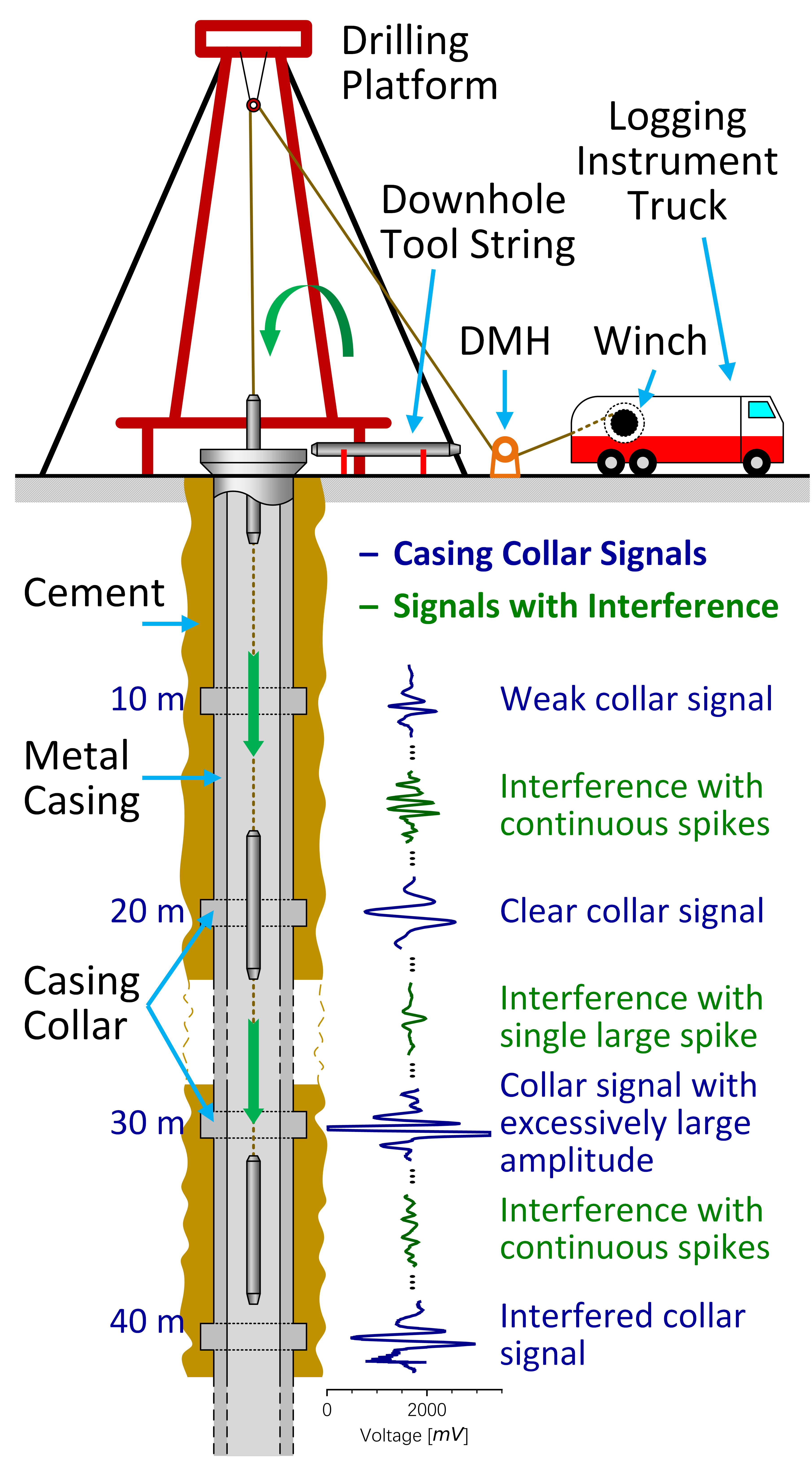}
    \caption{Cross-sectional illustration of a typical oil and gas well structure. Representative casing collar signatures from magnetic response are illustrated in dark blue near the corresponding casing collar, while typical interference signals are illustrated in dark green. Organized from \cite{xiao2025realization}.}
    \label{fig1}
\end{figure}

\IEEEPARstart{A}{ccurately} positioning downhole toolstrings (including perforating guns, bridge plugs, and packers) is essential in modern oil and gas well operations, directly affecting maximum productivity and operational safety \cite{harris1966effect}. Central to this task is precise downhole depth measurement, a challenge compounded by the extreme geometries of wellbores, as shown in Fig.~\ref{fig1} organized from \cite{xiao2025realization}, which often span thousands of meters in length while maintaining diameters of only a few inches \cite{seren2022miniaturized}.

While surface wheel measurement (SWM) method offers a cost-effective means of estimating depth via a depth measuring head (DMH) as the toolstring descends during wireline intervention operations, it is susceptible to errors induced by cable slippage and elastic stretch. Furthermore, SWM is inapplicable to emerging operations such as wireless perforating. Consequently, to achieve accurate depth measurement in the absence of a DMH, depth correction using a casing collar locator (CCL, a magnetic downhole positioning tool that detects magnetic anomalies at the casing collars of wellbores or pipes) is fundamental \cite{alvarez2018theory}. The CCL produces a characteristic magnetic response as an electrical signal when passing through each collar \cite{alvarez2018theory,gidado2023well,li2013casing}, termed a “CCL response” or “collar (magnetic) signature”. The characteristic magnetic response pattern typically exhibits a bipolar signature, as illustrated by dark blue waveforms in Fig.~\ref{fig1}. Through collar correlation, which refers to tying-in CCL logs with casing tally, depth reference markers are established, enabling actual and accurate depth measurement \cite{mijarez2014hpht,li2013casing}. The casing tally, also known as the list of collars or casing string reference depths, records the depths of casing collars and is commonly extracted from cementing quality data.

While collar correlation using CCL logs is an established method, collar signature recognition presents significant challenges. CCL signal integrity can be severely compromised by multiple factors, including cable effects, wellbore conditions \cite{brown1990effects}, toolstring motion (swing or rotation), amplifier saturation, and environmental noise \cite{mijarez2014hpht,wang2006application}. Consequently, collar signature waveforms become increasingly ambiguous, necessitating robust recognition of collar signatures amid various interferences, as illustrated by dark green waveforms in Fig.~\ref{fig1}.

Various signal processing techniques have been developed to identify collar signatures under interference conditions. Traditional methods include fixed or dynamic thresholding \cite{xiao2025realization,wang2012collardepth}, digital filters and template-based cross-correlation \cite{li2020application}, time-frequency domain techniques such as Fourier or wavelet transforms \cite{li2010approach}, and physical plausibility filters \cite{xiao2025realization,alvarez2018theory}. However, these approaches exhibit limited generalizability \cite{wang2006application,yang2025leak}. With the emergence of machine learning, researchers have increasingly employed deep neural networks to automate collar signature recognition. These developments include convolutional neural networks (CNNs) and long short-term memory networks (LSTMs) \cite{jing2025identification,ross2018generalized,le2016data}, Additionally, advanced architectures such as transformer models \cite{wen2022transformers} and physics-informed neural networks (PINNs) \cite{raissi2019physics} have been proposed for related fields such as downhole signal classification, anomaly detection, and denoising tasks \cite{noh2021deep,brazell2019machine,elhadidy2025optimizing,viggen2025improving,yang2025leak}.

Nevertheless, significant challenges persist. Neural networks require substantial volumes of labeled training data, which are often unavailable or difficult to obtain in downhole environments \cite{jing2025identification,murugan2017regularization}. This scarcity means that the available data for training collar recognition neural networks are considerably less than those for other tasks such as face recognition or image classification. Therefore, efficiently utilization of existing data is indispensable.

The limited sample amount and unique characteristics of CCL log data necessitate specialized preprocessing methods for neural network training. However, research on preprocessing methods remain scarce, despite extensive work on collar signature recognition \cite{alvarez2018theory,wang2006application,li2020application,li2010approach,yang2025leak,jing2025identification,wen2022transformers,noh2021deep,raissi2019physics}. Fortunately, extensive work exists on data augmentation for preventing overfitting and improving generalizability when training on small datasets. Data normalization methods -- including min-max scaling, Z-score normalization (standardization), and robust scaling -- demonstrate important roles in stabilizing input distributions and gradients, preventing vanishing and exploding gradients, and ultimately accelerating training convergence \cite{ioffe2015batch,santurkar2018does,bjorck2018understanding,asif2020effect}. Label smoothing regularization (LSR) and label distribution smoothing (LDS) discourage overconfident predictions, thereby enhancing model generalizability \cite{yang2021delving,szegedy2016rethinking}. Similarly, employing probability maps for boundary prediction instead of one-hot encoding (OHE) labels through boundary probabilization transforms the training objective from single-point prediction to probability distribution estimation, enabling models learn smooth and fuzzy decision boundaries that are more resistant to noise or perturbations \cite{yang2021delving,ross2018generalized,stoller2018wave}. Notably, Gaussian kernel often deliver optimal results in LDS applications \cite{yang2021delving}. Furthermore, graph augmentation techniques including randomly cropping, scaling, translation, and noise injection expand training datasets while improving model generalizability and robustness \cite{krizhevsky2017imagenet,krizhevsky2014one}.

The main contributions of this paper are as follows:
\begin{itemize}
\item We develop a system integrated into downhole toolstring, called the Signal Collecting Vessel (SCV), illustrated in Fig.~\ref{fig2}. The SCV samples raw CCL signals downhole and converts them to digital format, and stores them as waveforms for dataset construction.
\item We propose two neural networks models for collar signature recognition that serve as baselines for evaluating data preprocessing methods, as illustrated in Fig.~\ref{fig3}. The first model, Thin AlexNet (TAN), is modified from AlexNet -- a classic and proven architecture in pattern recognition. The second model, Miniaturized AlexNet (MAN), is a simplified version of TAN with fewer layers.
\item We propose several data augmentation methods for preprocessing original waveforms to enhance model training performance, including normalization, label distribution smoothing (LDS), label smoothing regularization (LSR), time scaling, cropping and translation, amplitude jittering, noise injection, and multiple sampling.
\item We conduct extensive experiments across various configuration combinations with filed CCL logs to validate our methods. Results demonstrate that standardization, LDS, and random cropping are fundamental requirements for models training, while LSR, time scaling, and multiple sampling significantly enhance model generalization capability.
\end{itemize}

\section{Methods}

\subsection{Problem Transformation}

As previously discussed, accurate downhole toolstring positioning via collar correlation relies on the correct identification of bipolar patterns (i.e., collar signatures) within CCL logs. The centroid of each bipolar pattern is typically designated as the instant the CCL coincides with a casing collar. In the absence of DMH assistance (specifically, without the depth indexing commonly utilized in wireline logging), time-series CCL logging becomes the requisite approach.

Field practice demonstrates that collar signatures are identifiable primarily through local waveform characteristics in the vicinity of the collar, rendering distant signal features negligible. Consequently, analyzing CCL log fragments via an appropriately sized sliding window is an effective strategy, aligning with methodologies proposed in \cite{ross2018generalized} and \cite{le2016data}. Furthermore, when raw CCL signals are sampled at a fixed frequency, absolute timestamps within the sliding windows become redundant; that is, the raw waveform sequence alone conveys sufficient information.

As shown in Fig.~\ref{fig1}, the casing tally, which provides a series of depths, correlates to the centroids of collar signatures (hereafter referred to as “collar marks”) on the temporal axis. The most direct representation of collar marks in CCL waveforms is one-hot encoding (OHE), where a value of 1 indicates the presence of a collar mark and 0 indicates its absence. This formulation frames collar mark prediction as a binary classification task.

However, collar marks exhibit extreme sparsity: Background samples (0s) outnumber target samples (1s) by several orders of magnitude, resulting in severe class imbalance. In backpropagation neural networks (BPNNs), this imbalance leads to sparse gradients and uneven penalty distribution, causing training instability and slow convergence. Consequently, training an effective classifier becomes computationally challenging. To mitigate this, OHE is replaced by a probability map \cite{geng2016label,ross2018generalized}, wherein labels represent the probability of boundary occurrence. Interestingly, consistent with the theory in \cite{geng2016label}, the label reflects the distribution of relative importance across possible categories (i.e., “the presence of a collar mark”). This transformation shifts the problem from hard binary classification to boundary membership estimation, providing denser feedback during backpropagation and facilitating stable and efficient network training.

In summary, the problem of collar signature recognition is transformed into a boundary membership estimation task, where the input is a windowed temporal CCL waveform and the output is a temporal probability map.

\subsection{Acquisition of Raw CCL Waveforms}

\begin{figure}[!b]
    \centering
    \includegraphics[width=1\linewidth]{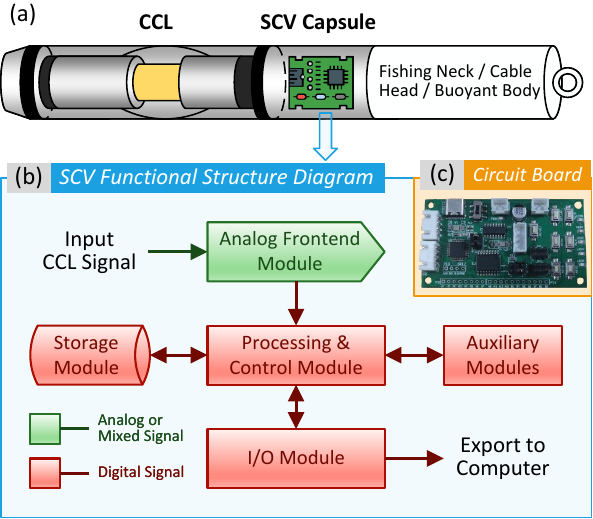}
    \caption{
        Structure of downhole toolstring integrated with the Signal Collection Vehicle (SCV):
        (a)~Schematic diagram of the internal structure of the perforating gun employed in this work;
        (b)~Functional structure diagram of the SCV;
        (c)~A SCV circuit board.
    }
    \label{fig2}
\end{figure}

To eliminate signal degradation associated with long cable transmission and preserve signal integrity, a Signal Collection Vehicle (SCV) was developed, as shown in Fig.~\ref{fig2}. The SCV is designed to be integrated into downhole toolstring and enable the real-time logging of raw CCL signals downhole.

The SCV comprises of the analog frontend (AFE) module, the signal processing and control module, the data storage module, the input-output (I/O) module, and auxiliary modules. As the SCV is lowered with downhole toolstring, the AFE module samples raw CCL signal at a sampling rate of \SI{1}{\kilo\hertz} via a 16-bit resolution analog-to-digital converter (ADC). The resulting digital data is recorded in the storage module. Upon completion of downhole operations, the SCV is salvaged, and data is exported via the I/O module.

The raw temporal CCL logs acquired by the SCV are designated as “original CCL waveforms”. Typically, a single log contains approximately 50 to 200 collar signatures, corresponding to a well depth ranging from \SI{500}{\meter} to \SI{2}{\kilo\meter}.

\subsection{Dataset Construction and Augmentation}

Collar marks within the field-acquired original CCL waveforms were manually annotated through expert analysis. To construct the dataset, waveforms surrounding each collar mark were fragmented into fixed-length fragments, while non-informative sections distant from the collars were excluded. As illustrated in Fig.~\ref{fig3}(a), each fragment is centered on a collar mark, which is initially represented using one-hot encoding.

To enhance training effectiveness, maximize data utilization, and avoid overfitting, we present various data augmentation methods for preprocessing original CCL data. These methods encompass data normalization, regularization, transformation and multiple sampling, which can be applied independently or in combination, as illustrated in Fig.~\ref{fig3}(b).

After preprocessing, the augmented waveform fragments and target labels constitute the datasets used for model training and evaluation.

\subsubsection{Normalization of Waveforms}~\nopagebreak

The original CCL waveform represents raw data from the ADC in unsigned integer format. Theoretically, raw data requires normalization to enhance training performance, particularly convergence speed \cite{lecun2002efficient}. Both min-max scaling and z-score normalization warrant investigation. For min-max scaling, waveforms are transformed to either $[0,1]$ or $[-1,1]$. The minimum and maximum values derive from either the waveform’s dynamic range or the ADC specifications.

\subsubsection{Label Distribution Smoothing (LDS)}~\nopagebreak

Convolving a kernel with the empirical density distribution produces a kernel-smoothed version. The effective label density distribution is defined as \cite{yang2021delving}:
\begin{equation}
    p^{\prime}(y^{\prime})\triangleq\int_{\mathcal{Y}}k(y,y^{\prime})p(y)\,dy
    \label{EQ1}
\end{equation}
where $p(y)$ is the label of $y$ in the training data; $p^{\prime}$ is the effective density of label $y^{\prime}$; and $k(y,y^{\prime})$ is the kernel function.

One-hot encoding of collars along the timeline constitutes a validity probability distribution that proves challenging for training. Convolution smooths these hard labels, enabling each label in the validity probability distribution to incorporate information from neighboring labels \cite{yang2021delving}.

While Gaussian kernels reportedly achieve optimal results among all kernel types \cite{yang2021delving}, other research cautions that Gaussian assumptions may not accommodate complex real-world datasets \cite{geng2016label}. This work employs a Gaussian kernel is employed, yielding the following label formulation:
\begin{equation}
    p^{\prime}(t)=
    \begin{cases}
        \sum\limits_{i}\sum\limits_{t}e^{-\frac{(t-t_i)^2}{2\sigma^2}} &, \quad if<1 \\\\
        1 &, \quad otherwise
    \end{cases}
    \label{EQ2}
\end{equation}
where $\sigma$ is the Gaussian root mean square (RMS) width; $t_i$ is the moment when the i\textsuperscript{th} collar occurs.

\subsubsection{Label Smoothing Regularization (LSR)}~\nopagebreak

LSR softens labels by redistributing a small probability portion from the correct class evenly among all classes. The distribution relationship follows \cite{szegedy2016rethinking}:
\begin{equation}
\begin{aligned}
    p(k) &= \delta_{k,i} \\
    p^{\prime}(k) &= (1-\epsilon)\delta_{k,i}+\frac{\epsilon}{K} \\
    \delta_{k,i} &=
    \begin{cases}
        1 &, \quad k=i \\
        0 &, \quad k\neq i
    \end{cases}
\end{aligned}
\label{EQ3}
\end{equation}
where $p(k)$ is the ground-truth distribution over the k\textsuperscript{th} class; $p^{\prime}(k)$ is the training distribution over the k\textsuperscript{th} class; $i$ denotes the correct class; $K$ is the total number of classes; $\delta$ is the Kronecker delta function; and $\epsilon$ is the smoothing parameter.

When employing the sigmoid function, the parameter $\epsilon$ effectively constrains the magnitude of the output logits. This mechanism discourages overconfident predictions by preventing the largest logit from becoming disproportionately larger than others, thereby regularizing the model and enhancing its generalization capability \cite{szegedy2016rethinking}.
In this study, given $K=2$, the training distribution simplifies to:
\begin{equation}
    p^{\prime}(k)=(1-\epsilon)p(k)+\frac{\epsilon}{2} \label{EQ4}
\end{equation}
where an $\epsilon$ value of 0.1 is recommended to strike a balance between accuracy and generalization capability.

\subsubsection{Geometric Transformations}~\nopagebreak

\begin{figure*}[p]
    \centering
    \includegraphics[width=\linewidth]{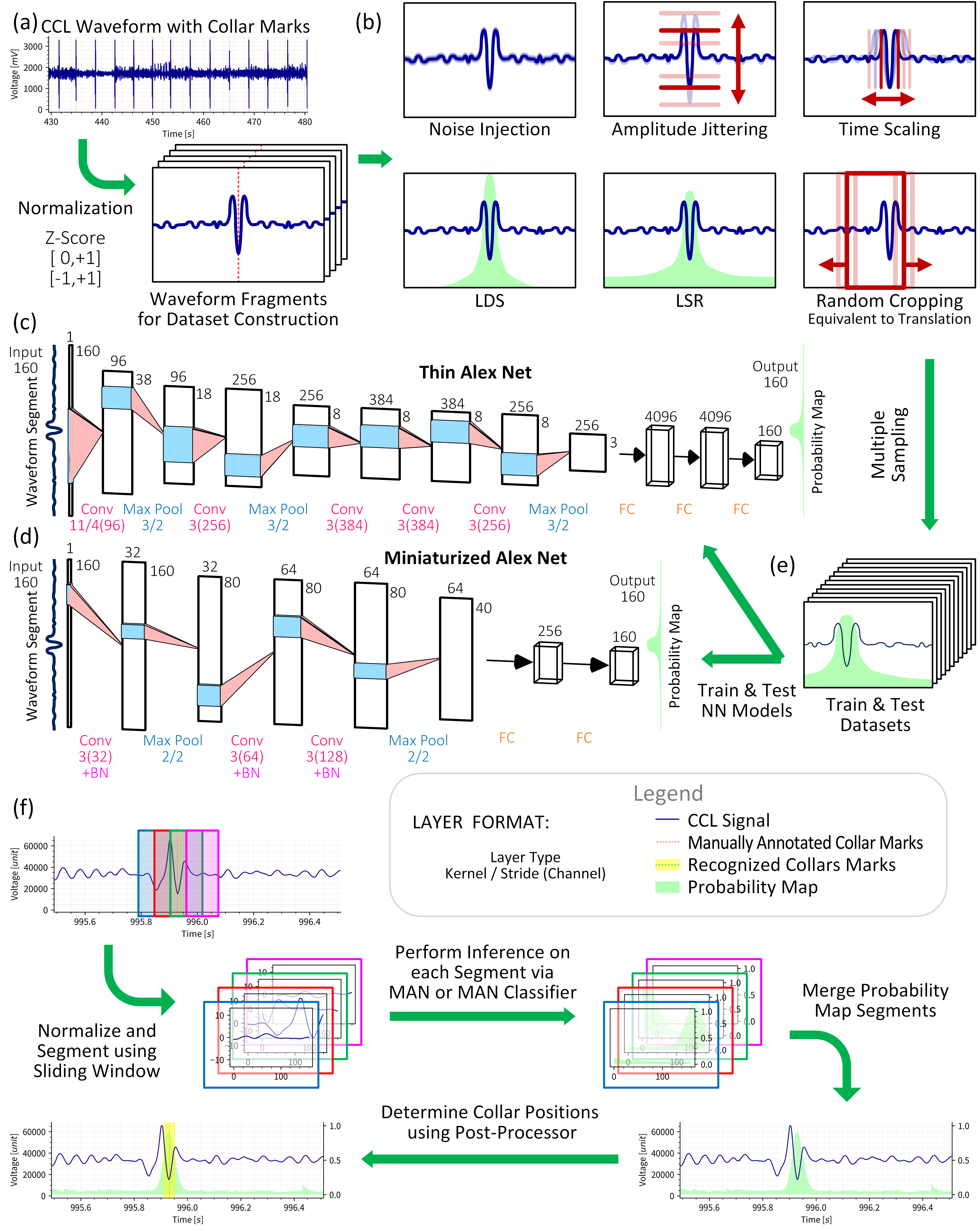}
    \caption{
        The train and inference process of this work:
        (a)~Fragment the normalized CCL waveform based on manually labeled collar marks, each fragment containing a collar mark at its center (waveform shapes in boxes are illustrative only);
        (b)~Augmentation methods for preprocessing waveform fragments and their labels;
        (c)--(d)~Baseline neural network architectures employed in this work;
        (e)~Multiple random augmentations applied to each fragment to generate diverse sub-samples variants for training and testing datasets;
        (f)~The procedure of casing collar recognition from CCL waveforms using sliding windows with overlap.
    }
    \label{fig3}
\end{figure*}

The dataset undergoes expansion through various geometric transformations, as illustrated in Fig.~\ref{fig3}(b). The following transformations are employed:
\begin{itemize}
\item \textbf{Time Scaling}: Waveform fragments are scaled along the time axis by random factors and subsequently resampled to restore the original sampling rate. The resampling process employs Hann-windowed sinc interpolation, the default resampling method in the TorchAudio library, to mitigate spectral artifacts, including ringing and aliasing, while ensuring effective high-frequency attenuation.
\item \textbf{Randomly Cropping and Translation}: These transformations waveform fragments into sub-samples that match both the sliding window length and the neural network model’s input length.
\item \textbf{Amplitude Jittering}: Waveform fragments are multiplied by random gain factors to enhance the model’s robustness and generalization capability.
\item \textbf{Noise Injection}: Gaussian noise is added to original fragments to improve model robustness against noise.
\item \textbf{Flipping}: Voltage or time axis flipping is excluded as such transformations would violate the physical principles governing CCL magnetic response.
\end{itemize}

\subsubsection{Multiple Sampling}~\nopagebreak

To maximize dataset utilization, each fragments undergoes multiple random augmentation process to generate diverse sub-samples variants, as illustrated in Fig.~\ref{fig3}(e). This approach theoretically accelerates convergence and enhance augmentation effectiveness.

\subsection{Neural Network}

The proposed model, Thin AlexNet (TAN), is a time-series version of the classical AlexNet architecture. The input and output dimensions are reduced from 2D to 1D to adapt CCL waveforms. Since AlexNet has demonstrated success in image classification \cite{krizhevsky2017imagenet,krizhevsky2014one} and possesses sufficient simplicity to clearly demonstrate the effects of data augmentation, the TAN model serves as an appropriate baseline.

The TAN model comprises 5 convolutional layers with ReLU activation, 3 max pooling layers, and 3 fully connected (FC) layers, with ReLU activation applied to the first two fully connected layers, as illustrated in Fig.~\ref{fig3}(c). The input of model is a fixed-length segment from a waveform. And the output produces a series of logits representing scores of collar mark classification for each temporal position within the input segments. This differs from the original AlexNet. Probability for each temporal position is obtained through sigmoid function.

Based on TAN model, we proposed a simplified version, Miniaturized AlexNet (MAN), as illustrated in Fig.~\ref{fig3}(d). MAN contains fewer convolutional layers, max pooling layers and fully connected layers than TAN. Additionally, batch normalization (BN) layers are appended in MAN to improve training stability. MAN employs the same input-output format as TAN.

\section{Experiments and Results}

\subsection{Evaluation Measures}

To evaluate the classifiers implemented by TAN and MAN, which output probability distributions for classification, metrics that measure distance and similarity between predicted and label distributions are appropriate \cite{geng2016label}. F1 score and cross-entropy (CE) represent standard choices for such evaluation. Additionally, the area under the precision-recall curve (AUC-PR) provides objective performance assessment for binary classifiers.

To recognize collar signatures from normalized long CCL waveforms, a sliding window of width $W$ with a strid of $\lfloor W/2\rfloor$ is employed. This configuration produces 50\% overlap between consecutive windows, ensuring that each overlapping region is captured by exactly two adjacent windows. The window advances through the waveform in half-width increments, balancing computational efficiency with temporal resolution while maintaining analytical continuity. The probability map segments inferenced from overlapping regions of adjacent windows are averaged to produce the complete probability map, as illustrated in Fig.~\ref{fig3}(f).

Through sliding window progression, the complete probability map for the entire CCL waveform is generated. Continuous intervals with probabilities exceeding the threshold are identified as valid regions, with their center positions designated as collar marks. This procedure constitutes the “post-processing” stage. The complete casing collar recognition process from waveform comprises a neural network-based classifier and a post-processor for probability map analysis, as illustrated in Fig.~\ref{fig3}(f). Collar recognition performance is evaluated by comparing recognized collar positions with manually annotated reference positions. Recognized collars within the neighborhood of annotated collars are classified as true positives, while those elsewhere are classified as false positives. Missed collars are considered as false negatives. Precision, recall and F1 score are calculated for evaluation.

\subsection{Training and Validation}

The CCL waveforms utilized in the experiments were acquired from field operations in Sichuan Province, China, ensuring that the results are representative of actual downhole conditions. The dataset was partitioned into training and validation subsets using a 3:1 ratio based on the CCL logs, yielding 288 and 50 original waveform fragments, respectively. To mitigate the limitations (including funds) imposed by the restricted dataset size, a multiple sampling strategy was employed to augment the training set. The training and validation processes were conducted offline on a workstation using the PyTorch framework.

\begin{figure*}[p]
    \centering
    \includegraphics[width=.9\linewidth]{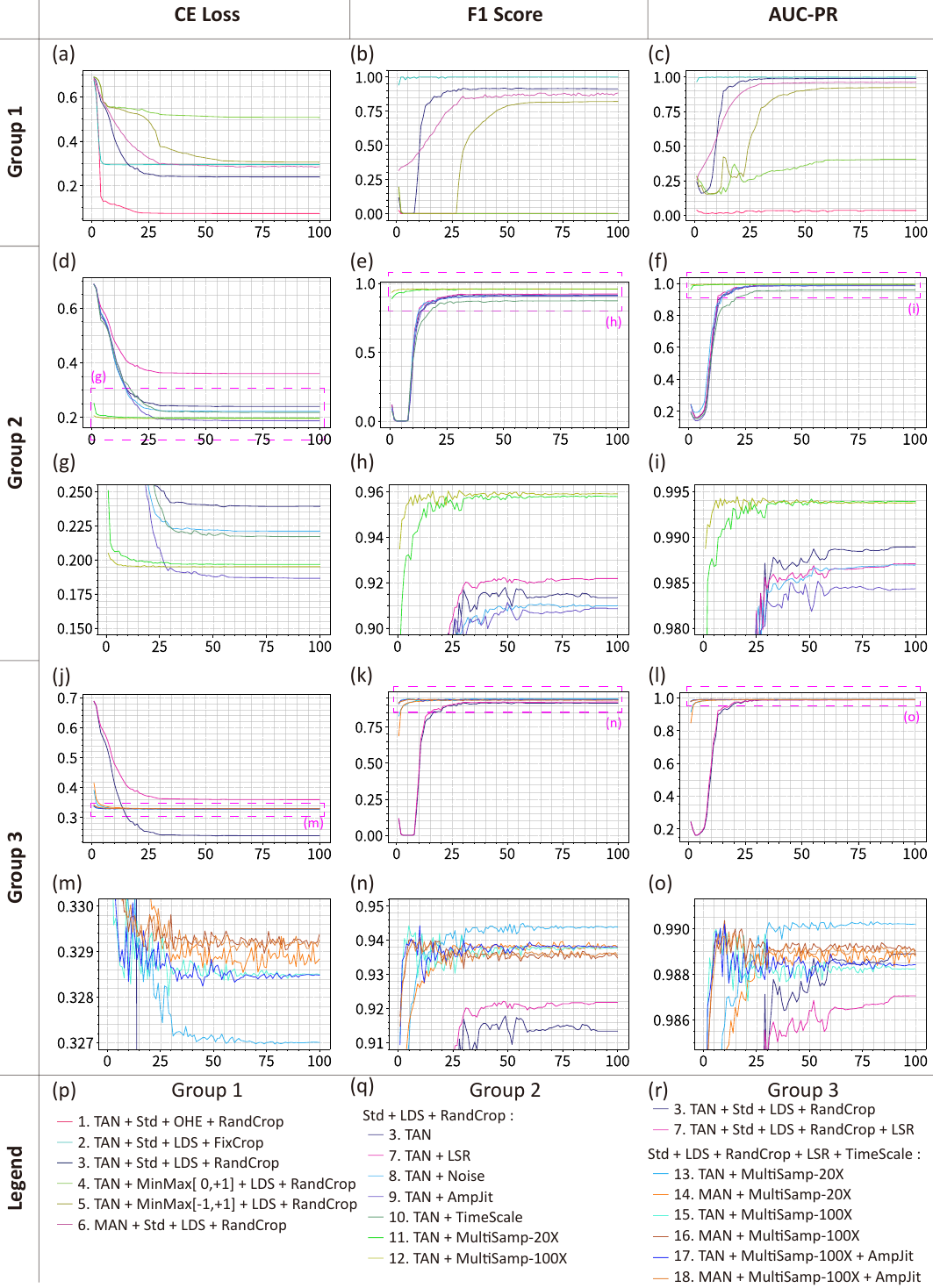}
    \caption{
        Evaluation metrics of training progress under different configurations, including cross-entropy loss, F1 score, and area under the precision-recall curve (AUC-PR).
        The curve indexes correspond to configurations in Tables~\ref{tab1} and \ref{tab2}. The meanings of abbreviations refer to Tables~\ref{tab1} and \ref{tab2}.
        (a)--(c)~Evaluation metrics for Group 1 configurations;
        (d)--(f)~Evaluation metrics for Group 2 configurations;
        (g)--(i)~Enlarged sections of (d)--(f);
        (j)--(l)~Evaluation metrics for Group 3 configurations;
        (m)--(o)~Enlarged sections of (j)--(l);
        (p)--(r)~legends for (a)--(c), (d)--(i), and (j)--(o), respectively.
    }
    \label{fig4}
\end{figure*}

\renewcommand{\arraystretch}{1.6}
\setlength{\tabcolsep}{3pt}
\begin{table*}[!tb]
\centering
\caption{
Experimental Results for Various Combinations of Normalization, Label Distribution, and Cropping Methods
}
\label{tab1}
\small

\resizebox{\textwidth}{!}{%
\begin{tabular}{@{} c | c c c c | l l l | l l l | l l l @{}}
\Xhline{1.2pt}
\multirow{2}{*}{\makecell[c]{\textbf{Cfg.}\\\textbf{No.}}} &
\multirow{2}{*}{\makecell[c]{\textbf{Model}}} &
\multirow{2}{*}{\makecell[c]{\textbf{Normalization}}} &
\multirow{2}{*}{\makecell[c]{\textbf{Lbl.}\\\textbf{Dis.}}} &
\multirow{2}{*}{\makecell[c]{\textbf{Crop}}} &
\multicolumn{3}{c|}{\makecell{Evaluation by Validation\\Set during Train}} &
\multicolumn{3}{c|}{\makecell{Evaluation by Moderate\\Interference Waveform}} &
\multicolumn{3}{c}{\makecell{Evaluation by Mild\\Interference Waveform}} \\
\cline{6-8}\cline{9-11}\cline{12-14}
& & & & &
\multicolumn{1}{c}{\textbf{CE}} &
\multicolumn{1}{c}{\textbf{F1}} &
\multicolumn{1}{c|}{\textbf{AUC-PR}} &
\multicolumn{1}{c}{\textbf{P}} &
\multicolumn{1}{c}{\textbf{R}} &
\multicolumn{1}{c|}{\textbf{F1}} &
\multicolumn{1}{c}{\textbf{P}} &
\multicolumn{1}{c}{\textbf{R}} &
\multicolumn{1}{c}{\textbf{F1}} \\
\Xhline{1.2pt}
\textbf{\textcolor{col1}{1}} & TAN & Standardization & OHE & Rand  & 0.0736 & 0      & 0.0350 & 0      & 0      & 0      & 0      & 0     & 0      \\
\textbf{\textcolor{col2}{2}} & TAN & Standardization & LDS & Fix   & 0.2943 & 1      & 0.9987 & 0      & 0      & 0      & 0      & 0     & 0      \\
\textbf{\textcolor{col3}{3}} & TAN & Standardization & LDS & Rand  & 0.2391 & 0.9134 & 0.9889 & 0.9136 & 0.9610 & 0.9367 & 0.9811 & 1     & 0.9905 \\
\textbf{\textcolor{col4}{4}} & TAN & MinMax [ 0,+1]  & LDS & Rand  & 0.5074 & 0      & 0.4051 & 0      & 0      & 0      & 0      & 0     & 0      \\
\textbf{\textcolor{col5}{5}} & TAN & MinMax [-1,+1]  & LDS & Rand  & 0.3055 & 0.8205 & 0.9234 & 0.0396 & 0.1169 & 0.0592 & 0.0335 & 0.1346& 0.0536 \\
\textbf{\textcolor{col6}{6}} & MAN & Standardization & LDS & Rand  & 0.2852 & 0.8821 & 0.9619 & 1      & 0.9091 & 0.9524 & 1      & 1     & 1      \\
\Xhline{1.2pt}
\end{tabular}%
} 
\vspace{1ex}

\begin{minipage}{\textwidth}
\footnotesize
All configurations use a batch size of 16, with no additional data augmentation methods applied.\\
Abbreviations: Cfg. No.~=~configuration number; Lbl. Dis.~=~label distribution; CE~=~cross-entropy; F1~=~F1 Score; AUC-PR~=~area under the precision-recall curve; P~=~precision; R~=~recall; OHE~=~one-hot encoding; LDS~=~label distribution smoothing; Rand~=~random.\\
Configuration numbers are color-coded to match the corresponding curves in Fig.~\ref{fig4}.
\end{minipage}

\end{table*}

\begin{table*}[!tb]
\centering
\caption{
Experimental Results for Various Combinations of Soft Label, \\
Geometric Transformations, and Multiple Sampling Methods
}
\label{tab2}
\small

\resizebox{\textwidth}{!}{%
\begin{tabular}{@{} c | c c c c c c | l l l | l l l | l l l @{}}
\specialrule{1.2pt}{0pt}{0pt}
\multirow{2}{*}{\makecell[c]{\textbf{Cfg.}\\\textbf{No.}}} &
\multirow{2}{*}{\makecell[c]{\textbf{Model}}} &
\multirow{2}{*}{\makecell[c]{\textbf{Soft}\\\textbf{Label}}} &
\multirow{2}{*}{\makecell[c]{\textbf{Noise}\\\textbf{Inj.}}} &
\multirow{2}{*}{\makecell[c]{\textbf{Amp.}\\\textbf{Jit.}}} &
\multirow{2}{*}{\makecell[c]{\textbf{Time}\\\textbf{Scale}}} &
\multirow{2}{*}{\makecell[c]{\textbf{Multi.}\\\textbf{Samp.}}} &
\multicolumn{3}{c|}{\makecell{Evaluation by Validation\\Set during Train}} &
\multicolumn{3}{c|}{\makecell{Evaluation by Moderate\\Interference Waveform}} &
\multicolumn{3}{c }{\makecell{Evaluation by Mild\\Interference Waveform}} \\
\cline{8-10}\cline{11-13}\cline{14-16}
& & & & & & &
\multicolumn{1}{c }{\textbf{CE}} &
\multicolumn{1}{c }{\textbf{F1}} &
\multicolumn{1}{c|}{\textbf{AUC-PR}} &
\multicolumn{1}{c }{\textbf{P}} &
\multicolumn{1}{c }{\textbf{R}} &
\multicolumn{1}{c|}{\textbf{F1}} &
\multicolumn{1}{c }{\textbf{P}} &
\multicolumn{1}{c }{\textbf{R}} &
\multicolumn{1}{c }{\textbf{F1}} \\
\specialrule{1.2pt}{0pt}{0pt}
\textbf{\textcolor{col3}{3}}   & TAN & -   & - & - & - & 1   & 0.2391 & 0.9134 & 0.9889 & 0.9136 & 0.9610 & 0.9367 & 0.9811 & 1      & 0.9905 \\
\textbf{\textcolor{col6}{6}}   & MAN & -   & - & - & - & 1   & 0.2852 & 0.8821 & 0.9619 & 1      & 0.9091 & 0.9524 & 1      & 1      & 1      \\
\hline
\textbf{\textcolor{col7}{7}}   & TAN & LSR & - & - & - & 1   & 0.3596 & 0.9217 & 0.9871 & 0.9610 & 0.9610 & 0.9610 & 0.9811 & 1      & 0.9905 \\
\textbf{\textcolor{col8}{8}}   & TAN & -   & + & - & - & 1   & 0.2209 & 0.9098 & 0.9869 & 0.8152 & 0.9740 & 0.8876 & 0.9630 & 1      & 0.9811 \\
\textbf{\textcolor{col9}{9}}   & TAN & -   & - & + & - & 1   & 0.1864 & 0.9085 & 0.9843 & 0.9367 & 0.9610 & 0.9487 & 0.9811 & 1      & 0.9905 \\
\textbf{\textcolor{col10}{10}} & TAN & -   & - & - & + & 1   & 0.2170 & 0.8726 & 0.9592 & 0.9740 & 0.9740 & 0.9740 & 1      & 1      & 1      \\
\textbf{\textcolor{col11}{11}} & TAN & -   & - & - & - & 20  & 0.1966 & 0.9577 & 0.9939 & 0.9390 & 1      & 0.9686 & 1      & 1      & 1      \\
\textbf{\textcolor{col12}{12}} & TAN & -   & - & - & - & 100 & 0.1949 & 0.9589 & 0.9937 & 0.9872 & 1      & 0.9935 & 1      & 1      & 1      \\
\hline
\textbf{\textcolor{col13}{13}} & TAN & LSR & - & - & - & 20  & 0.3270 & 0.9439 & 0.9902 & 0.9506 & 1      & 0.9747 & 1      & 1      & 1      \\
\textbf{\textcolor{col14}{14}} & MAN & LSR & - & - & - & 20  & 0.3288 & 0.9379 & 0.9888 & 1      & 1      & 1      & 0.9811 & 1      & 0.9905 \\
\textbf{\textcolor{col15}{15}} & TAN & LSR & - & - & - & 100 & 0.3285 & 0.9377 & 0.9883 & 0.9625 & 1      & 0.9809 & 1      & 1      & 1      \\
\textbf{\textcolor{col16}{16}} & MAN & LSR & - & - & - & 100 & 0.3294 & 0.9348 & 0.9890 & 0.9744 & 0.9870 & 0.9806 & 0.9808 & 0.9808 & 0.9808 \\
\textbf{\textcolor{col17}{17}} & TAN & LSR & - & + & - & 100 & 0.3285 & 0.9381 & 0.9884 & 0.9620 & 0.9870 & 0.9744 & 0.9811 & 1      & 0.9905 \\
\textbf{\textcolor{col18}{18}} & MAN & LSR & - & + & - & 100 & 0.3293 & 0.9355 & 0.9889 & 0.9744 & 0.9870 & 0.9806 & 0.9808 & 0.9808 & 0.9808 \\
\specialrule{1.2pt}{0pt}{0pt}
\end{tabular}
}

\vspace{1ex}
\begin{minipage}{\linewidth}
\footnotesize
All configurations employ standardization for normalization, LDS for label distribution, random cropping, and a batch size of 16. \\
Additional abbreviations: Inj.~=~injection; Amp. Jit.~=~amplitude jittering; Multi. Samp.~=~multiple sampling; LSR~=~label smoothing regularization.
\end{minipage}

\end{table*}

The models are trained with a batch size of 16 for 100 epochs utilizing cross-entropy loss (CE loss) with an adaptive moment estimation (Adam) optimizer. Key configurations (Cfgs.) are presented in Tables~\ref{tab1} and \ref{tab2}. The training progress and the optimal model performance for each configuration are illustrated in Fig.~\ref{fig4}. Additionally, model inference and post-processing were evaluated on two full-length waveforms characterized by mild and moderate interference, respectively. The mild interference waveform contains 52 collars with maximum depth of \SI{512.2}{\meter}, while the moderate interference waveform contains 77 collars with maximum depth of \SI{771.1}{\meter}. Crucially, the waveform fragments from these evaluation waveforms were excluded from the training set. The precision, recall, and F1 score results are tabulated in Tables~\ref{tab1} and \ref{tab2}, with the aggregated results for both waveforms are tabulated in Table~\ref{tab3}.

\subsection{Results and Analysis}

\begin{figure*}[!b]
    \centering
    \includegraphics[width=1\linewidth]{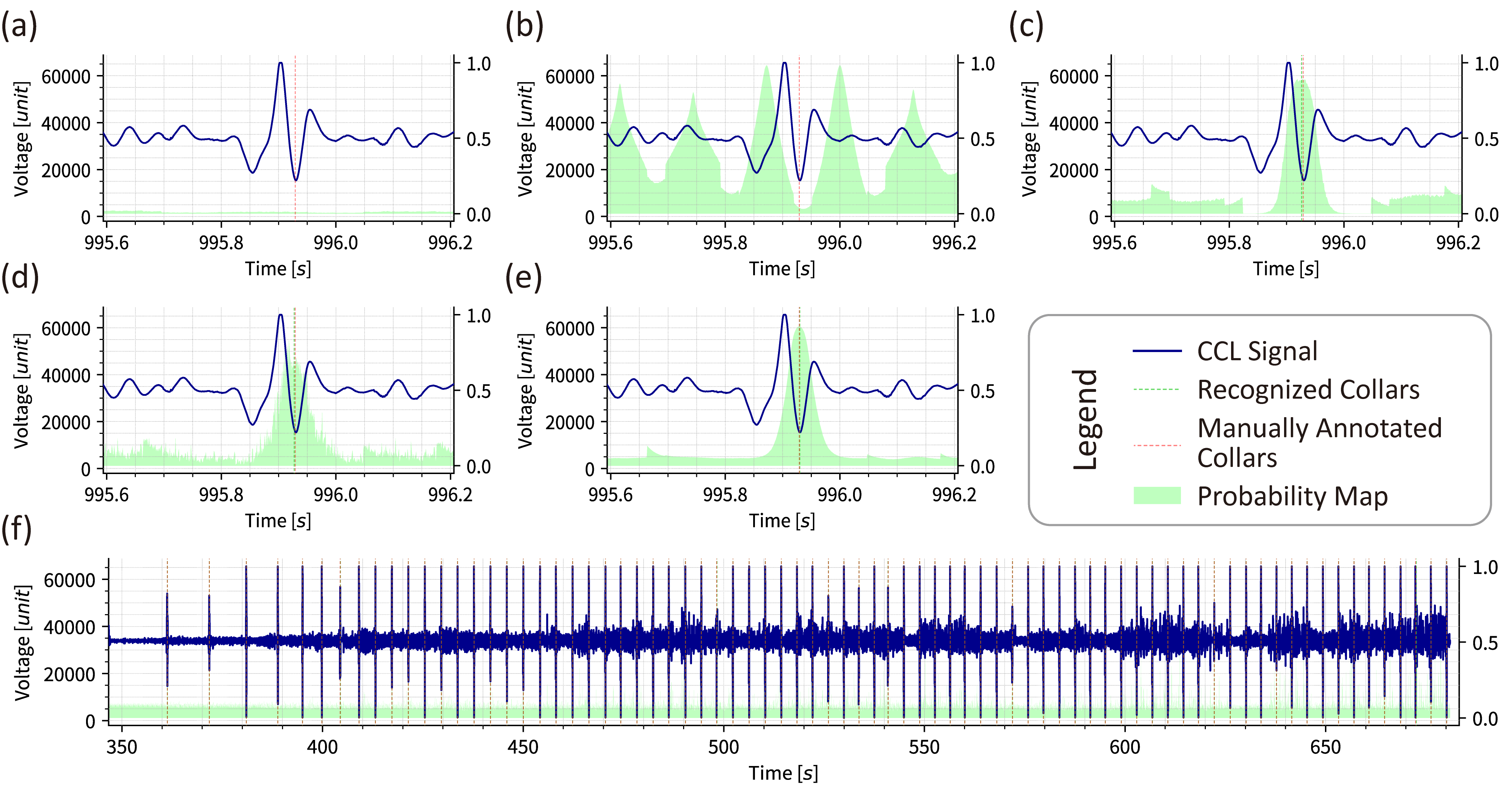}
    \caption{
        Probability maps and recognition results for various configurations.
        (a)~Cfg.~1 (using OHE);
        (b)~Cfg.~2 (using fixed cropping);
        (c)~Cfg.~3 (using LDS and random cropping);
        (d)~Cfg.~6 (using fundamental methods with MAN model);
        (e)~Cfg.~13 (optimal combination candidate using TAN model);
        (f)~Full results of Cfg.~14 (optimal combination candidate using MAN model).
    }
    \label{fig5}
\end{figure*}

\newcommand{\cdash}{        
    \makecell[c]{--}
}

\begin{table*}[!b]
    \centering
    \begin{minipage}{0.9\textwidth}
        \caption{
            Comparison of Performance with Other Works
        }
        \label{tab3}
        \setlength{\tabcolsep}{6pt}
        \resizebox{\textwidth}{!}{
            \begin{tabular}{c | rllll | l}
\specialrule{1.2pt}{0pt}{0pt}
    \multicolumn{1}{c|}{\multirow{2}{*}{\textbf{Network}}}     &
    \multicolumn{5}{c|}{\multirow{1}{*}{\textbf{Performance}}} &
    \multicolumn{1}{c }{\multirow{2}{*}{\textbf{Method}}}      \\
    \cline{2-6}
    ~                                  &
    \multicolumn{1}{c}{\textbf{Tests}} &
    \multicolumn{1}{c}{\textbf{Acc}}   &
    \multicolumn{1}{c}{\textbf{P}}     &
    \multicolumn{1}{c}{\textbf{R}}     &
    \multicolumn{1}{c|}{\textbf{F1}}   &
    ~                                  \\
\specialrule{1.2pt}{0pt}{0pt}
    {\textcolor{col3} {Cfg.~3} }    & 129    & 0.920  & 0.940  & 0.977  & 0.958  & CCL + TAN + LDS                                           \\
    {\textcolor{col6} {Cfg.~6} }    & 129    & 0.946  & 1      & 0.946  & 0.972  & CCL + MAN + LDS                                           \\
    {\textcolor{col13}{Cfg.~13}}    & 129    & 0.970  & 1      & 0.970  & 0.985  & CCL + TAN + LDS + Data Augmentation                       \\
    {\textcolor{col14}{Cfg.~14}}    & 129    & 0.992  & 1      & 0.992  & 0.996  & CCL + MAN + LDS + Data Augmentation                       \\
    {\cite{jing2025identification}} & 269    & 0.974  & 1      & 0.942  & 0.970  & CCL + CNN                                                 \\
    {\cite{jing2025identification}} & 269    & 0.948  & 1      & 0.884  & 0.939  & CCL + LSTM                                                \\
    {\cite{jing2025identification}} & 269    & 0.978  & 0.959  & 0.991  & 0.975  & CCL + CNN-LSTM                                            \\
    {\cite{xiao2025realization}}    & 579    & 0.973  & 0.988  & 0.985  & 0.986  & CCL + Dynamic amplitude threshold + Physical plausibility \\
    {\cite{wang2012collardepth}}    & 8      & 1      & \cdash & \cdash & \cdash & CCL + Relative amplitude                                  \\
    {\cite{mijarez2014hpht}}        & \cdash & \cdash & \cdash & \cdash & \cdash & CCL + Cross correlation + Predifined threshold            \\
    {\cite{li2013casing}}           & \cdash & \cdash & \cdash & \cdash & \cdash & CCL + Wavelet transform                                   \\
\specialrule{1.2pt}{0pt}{0pt}
\end{tabular}
        }
        \vspace{0.5ex}
        \small
        \noindent\\
        “--” indicates “Not mentioned”; “P”, “R” are precision and recall, respectively.
    \end{minipage}
\end{table*}

\let\cdash\undefined

To explore the validity and importance of augmentation methods, several real and complete waveforms are experimented to verify the models’ ability of predicting collar positions. The experimental results are indexed and divided into 3 groups, as illustrated in Fig.~\ref{fig4}, with curve indexes corresponding to those in Tables~\ref{tab1} and \ref{tab2}. The first group (Cfgs.~1--6) examines normalization methods, label distribution methods, and cropping methods. The second group (Cfgs.~7--12) investigates label smoothing, noise injection, amplitude jittering, and time scaling. The final group (Cfgs.~13--18) explores optimal combination configurations.

\subsubsection{Fundamental Preprocessing Requirements}~\nopagebreak

Comparison of Cfgs.~1 and 3, which differ in label distribution method, confirms that F1 score for both classifier and recognition using one-hot encoding approaches 0, despite the CE loss decreasing more rapidly than with LDS. The probability maps corroborate this observation, as illustrated in Figs.~\ref{fig5}(a,c). This indicates that one-hot encoded collar labels are difficult for models to learn, as the loss function reaches a local minimum when models consistently output negative predictions, as illustrated in Fig.~\ref{fig4}(a).

Similar phenomena occur with min-max normalization versus standardization (corresponding to Cfgs.~3--5), and fixed cropping versus random cropping (corresponding to Cfgs.~2 and 3), as illustrated in Figs.~\ref{fig5}(b,c). Waveforms processed with min-max normalization prove challenging for model training. Furthermore, models using fixed cropping learn only sliding window positions rather than waveform characteristics.

Comparison of Cfgs.~3 and 6 clearly demonstrates that both TAN and MAN module have the ability to estimate collar positions when using LDS, standardization, and random cropping, as illustrated in Figs.~\ref{fig5}(c,d). However, MAN training proceeds more slowly than TAN training, with slightly inferior performance, as MAN contains approximately half the parameters of TAN. These findings establish that LDS, standardization, and random cropping are fundamental requirements for collar recognition model training.

\subsubsection{Generalization Enhancement Methods}~\nopagebreak

To investigate the effects of LSR, noise injection, amplitude jittering, and time scaling, Cfgs.~7--10 based on Cfg.~3 are experimented. Comparison of curves in Figs.~\ref{fig4}(d--i) reveals that: (a)~the final CE loss for Cfgs.~8--10 is lower than that for control Cfg.~3, while Cfg.~7 shows higher loss; (b)~the classifier F1 scores of Cfgs.~7--9 are similar, while Cfg.~10 shows slightly lower performance; (c)~LSR convergence is slower than others. Performance comparison between Cfgs.~7--10 and control Cfg.~3, as tableted in Table~\ref{tab2}, demonstrates that: (a)~LSR exhibits higher CE loss but superior F1 score in waveform evaluation, with 0.024 improvement on the moderate inference waveform; (b)~although amplitude jittering and time scaling show lower F1 scores on validation set evaluation is compared to the control configuration, their waveform evaluation F1 scores are higher, with improvements of 0.012 and 0.037 respectively on the moderate inference waveform. These results suggest that LSR enhances generalization capability at the cost of convergence speed, while amplitude jittering and time scaling also improve generalization. However, noise injection performs worse than all other configurations. Additional experiments reveal that small noise provides limited benefits to performance, while large noise impairs performance. This likely occurs because real-world waveforms inherently contain small noise, and additional large noise hinders model learning waveform characteristics.

To investigate the effects of multiple sampling, Cfgs.~11--12 based on Cfg.~3 are experimented. Configurations with multiple sampling converge substantially faster than the control configuration, as shown in Figs.~\ref{fig4}(d--i), because multiple sampling increases iterations per epoch. The F1 score improves up to 0.045 in evaluation.

In summary, LSR, amplitude jittering, time scaling, and multiple sampling significantly enhance generalization capability, while noise injection provides limited benefits. By comparing Cfg.~13 and Cfg.~14 with Cfg.~3 and Cfg.~6, respectively, these data augmentation methods achieve F1 score improvements of up to 0.027 for TAN model and 0.024 for MAN model.

\subsubsection{Optimal Configuration Identification}~\nopagebreak

Based on these conclusions, Cfgs.~13--18 are experimented to identify optimal combinations while comparing with Cfgs.~3, 6 and 7, as illustrated in Figs.~\ref{fig4}(j--o) and ~\ref{fig5}(c--e) and tabulated in Table~\ref{tab2}. Several noteworthy phenomena emerge from these experiments.

First, metrics of configurations with $100\times$ multiple sampling fluctuate more dramatically than those with $20\times$, suggesting the need for smaller initial learning rates when applying extensive multiple sampling. Second, simultaneous use of amplitude jittering and time scaling unexpectedly reduces the performance. We hypothesize that amplitude scaling compromises standardization benefits while improving generalization, as the relative amplitude of waveforms contains critical information. Third, Cfg.~13 achieves the highest CE loss and validation set F1 score among all TAN model configurations, yet its F1 score on the moderate interference waveform does not show corresponding improvement, as illustrated in Figs.~\ref{fig4}(m--o). This occurs because multiple sampling introduces more randomly preprocessed samples, enhancing generalization capability.

Notably, both full-length evaluation waveforms contain several dozens of collar signatures, effectively functioning as a small-scale evaluation set. Consequently, it is plausible to achieve near-perfect or perfect results (an F1 score of 1.0) provided the model possesses sufficient accuracy. Furthermore, in the case of waveforms with mild interference, the collar signatures exhibit greater separability, thereby facilitating the attainment of perfect identification performance.

Most significantly, MAN performance does not degrade substantially compared to TAN, and MAN exhibits less performance degradation on moderate interference waveform than TAN, despite containing approximately half the parameters, as tabulated in Table~\ref{tab2}. These findings suggest that: (a)~TAN may experience overfitting in certain dimensions; (b)~collar classification could potentially be achieved with more compact networks warranting future investigation; (c)~practical applications can consider the trade-off between accuracy and parameter count.

\subsubsection{Performance Validation and Key Findings}~\nopagebreak

The results of Cfg.~14 are illustrated in Fig.~\ref{fig5}(f) as an example, while the details of other experiments are not repeated here. Casing collars recognized by MAN model and collars labeled manually are marked in green and red, respectively. The casing collar positions are correctly recognized and align completely with manual annotations, providing the essential foundation for precise depth measurement.

These findings establish that the standardization, LDS, and random cropping are fundamental preprocessing requirements for collar recognition model training, while LSR, time scaling, and multiple sampling significantly enhance model generalization capability. The data augmentation methods achieve F1 score improvements of up to 0.063 for TAN model waveform evaluation and up to 0.048 for MAN model evaluation, compared to configurations using only fundamental methods, representing notable performance enhancements. Furthermore, the baseline models TAN and MAN exhibit performance comparable to prior works, as tabulated in Table~\ref{tab3}. However, applying the proposed data augmentation methods yields F1 score improvements of up to 0.045 for TAN model and 0.057 for MAN model relative to other approaches. This observation demonstrates that the proposed data augmentation methods are effective even when the underlying model architecture is not optimal.

\section{Conclusion}

This work developed the SCV to acquire raw downhole CCL signals for dataset construction. The contributions of various data augmentation methods were systematically analyzed through baseline evaluations using the proposed neural network recognition methodology. Experimental validation with field CCL logs confirms the effectiveness and validity of proposed approaches. Results indicate that standardization, LDS, and random cropping are fundamental preprocessing prerequisites for training collar recognition models, while LSR, time scaling, and multiple sampling significantly enhance model generalization capabilities. Incorporating the proposed augmentation methods into the two baseline models results in maximum F1 score improvements of 0.027 and 0.024 for the TAN and MAN models, respectively. Furthermore, applying these techniques yields F1 score gains of up to 0.045 for the TAN model and 0.057 for the MAN model compared to prior studies. This research addresses the existing gaps in data augmentation methodologies for training casing collar recognition models under CCL data-limited conditions, and provides a technical foundation for the future automation of downhole operations.

\bibliographystyle{IEEEtran}
\bibliography{reference}

@article{xiao2025realization,
  title         = {Realization of Precise Perforating Using Dynamic Threshold and Physical Plausibility Algorithm for Self-Locating Perforating in Oil and Gas Wells},
  author        = {Xiao, Si-Yu and Ren, Guo-Hui and Mao, Tian-Hao and Chen, Yu-Qiao and Liu, Yi-An and Wang, Jun-Jie and Tang, Kai and Zhao, Xin-Di and Yu, Zhi-Jian and Liu, Shuang and Chen, Tu-Pei and Yang, Liu},
  journal       = {arXiv preprint arXiv:2509.00608},
  year          = {2025},
  archiveprefix = {arXiv},
  primaryclass  = {eess.SY},
  doi           = {10.48550/arXiv.2509.00608}
}

@article{harris1966effect,
  title     = {The Effect of Perforating Oil Well Productivity},
  author    = {Harris, M.H.},
  journal   = {Journal of Petroleum Technology},
  volume    = {18},
  number    = {04},
  pages     = {518--528},
  year      = {1966},
  month     = {04},
  _ublisher = {SPE},
  _address  = {Richardson, TX, USA},
  _issn     = {0149-2136},
  doi       = {10.2118/1236-PA}
}

@article{seren2022miniaturized,
  title     = {Miniaturized Casing Collar Locator for Small Downhole Robots},
  author    = {Seren, Huseyin Rahmi and Deffenbaugh, Max},
  journal   = {IEEE Sensors Letters},
  volume    = {6},
  number    = {4},
  pages     = {1--4},
  year      = {2022},
  _ublisher = {IEEE},
  _address  = {Piscataway, NJ, USA},
  doi       = {10.1109/LSENS.2022.3158002}
}

@article{alvarez2018theory,
  title     = {Theory, Design, Realization, and Field Results of An Inductive Casing Collar Locator},
  author    = {Alvarez, Jose Oliverio and Buzi, Erjola and Adams, Robert W and Deffenbaugh, Max},
  journal   = {IEEE Transactions on Instrumentation and Measurement},
  volume    = {67},
  number    = {4},
  pages     = {760--766},
  year      = {2018},
  _ublisher = {IEEE},
  _address  = {Piscataway, NJ, USA},
  doi       = {10.1109/TIM.2018.2795138}
}

@inproceedings{gidado2023well,
  title     = {Well Diagnostic of New Underperforming Wells Using Downhole Log Tool {[SNT \& MDT]}},
  author    = {Gidado, A. O. and Ekesiobi, C. and Kpone-Tonwe, H. and Adesun, J.},
  booktitle = {SPE Nigeria Annual International Conference and Exhibition},
  _location = {Lagos, Nigeria},
  pages     = {D021S012R001},
  year      = {2023},
  month     = {07},
  _ublisher = {SPE},
  _address  = {Richardson, TX, USA},
  doi       = {10.2118/217236-MS}
}

@inproceedings{li2013casing,
  title     = {Casing State Detection Methods Based on the {CCL} Signal of the Tractor for Horizontal Wells},
  author    = {Li, Haoyu and Tang, Tiantian and Wang, Yanjun},
  booktitle = {2013 IEEE 11th International Conference on Electronic Measurement \& Instruments},
  _location = {Harbin, China},
  volume    = {2},
  pages     = {568--573},
  year      = {2013},
  month     = {08},
  _ublisher = {IEEE},
  _address  = {Piscataway, NJ, USA},
  doi       = {10.1109/ICEMI.2013.6743143}
}

@inproceedings{mijarez2014hpht,
  title     = {{HPHT} Cased-Hole {CCL} Tool Enhancement via {DSP} Techniques for Accurate Depth Control in Wire-Line Well Interventions},
  author    = {Mijarez, Rito and Pascacio, David and Guevara, Ricardo and Tello, Carlos and Pacheco, Olimpia and Rodr{\'\i}guez, Joaqu{\'\i}n},
  booktitle = {International Conference on High Temperature Electronics},
  _location = {Albuquerque, NM, USA},
  pages     = {305--310},
  year      = {2014},
  month     = {05},
  _ublisher = {IMAPS},
  _address  = {Pittsburgh, PA, USA},
  doi       = {10.4071/HITEC-THA15}
}

@article{brown1990effects,
  title   = {The Effects of Cable on Signal Quality},
  author  = {Brown, Jim},
  journal = {Sound and Video Contractor},
  pages   = {22--33},
  year    = {1990}
}

@article{wang2006application,
  title   = {Application of Computer Automatic Discriminating Technology to the Depth Control of Perforation},
  author  = {Wang, Hong-Liang and Tang, Wen-Jiang},
  journal = {Well Logging Technology},
  volume  = {30},
  number  = {4},
  pages   = {378--380},
  year    = {2006},
  _issn   = {1004-1338},
  _doi    = {10.16489/j.issn.1004-1338.2006.04.027}
}

@article{wang2012collardepth,
  title   = {Study on Collar Depth Identification Based on Relative Amplitude Method},
  author  = {Wang, Hui and Lv, Haixia and Pan, Junhui and Li, Guojia and Gao, Xing},
  journal = {Journal of Harbin University of Commerce (Natural Sciences Edition)},
  volume  = {28},
  number  = {4},
  pages   = {435--438},
  year    = {2012},
  _issn   = {1672-0946},
  doi     = {10.19492/j.cnki.1672-0946.2012.04.016}
}

@article{li2020application,
  title   = {Application of Cross Correlation Function Method in Locating Perforation Depth},
  author  = {Li, Jin and Liu, Yuhai and Zhang, Jian and Wang, Jiang and Zhang, Yiling},
  journal = {Journal of Southwest Petroleum University (Natural Science Edition)},
  year    = {2020},
  volume  = {42},
  number  = {6},
  pages   = {42--48},
  _issn   = {1674-5086},
  _doi    = {10.11885/j.issn.1674-5086.2020.06.24.01}
}

@article{li2010approach,
  title   = {Study of the Approach to Extract Casing Collar Locator Information Features Based on Anti-aliasing Wavelet Time Entropy in Frequency Domain},
  author  = {Li, Haoyu and Chen, Jikai and Xiao, Yong and Liu, Xingbin and Wu, Jianqiang},
  journal = {High Technology Letters},
  year    = {2010},
  volume  = {20},
  number  = {5},
  pages   = {538--543},
  _issn   = {1002-0470},
  _doi    = {10.3772/j.issn.1002-0470.2010.05.016}
}

@article{yang2025leak,
  title     = {Leak Identification and Positioning Strategies for Downhole Tubing in Gas Wells},
  author    = {Yang, Yun Peng and Luan, Guo Hua and Zhang, Lian Fang and Niu, Ming Yong and Zou, Guang Gui and Zhang, Xu Liang and Wang, Jin You and Yang, Jing Feng and Li, Mo Song},
  journal   = {Processes},
  volume    = {13},
  number    = {6},
  pages     = {1708},
  year      = {2025},
  _ublisher = {MDPI},
  _address  = {Basel, Switzerland},
  _issn     = {2227-9717},
  doi       = {10.3390/pr13061708}
}

@article{jing2025identification,
  title     = {Identification and Prediction of Casing Collar Signal Based on {CNN-LSTM}},
  author    = {Jing, Jun and Qin, Yiman and Zhu, Xiaohua and Shan, Hongbin and Peng, Peng},
  journal   = {Arabian Journal for Science and Engineering},
  volume    = {50},
  number    = {7},
  pages     = {4897--4911},
  year      = {2025},
  _ublisher = {Springer},
  _address  = {Berlin, Germany},
  _issn     = {2191-4281},
  doi       = {10.1007/s13369-024-09440-5}
}

@article{ross2018generalized,
  title     = {Generalized seismic phase detection with deep learning},
  author    = {Ross, Zachary E and Meier, Men-Andrin and Hauksson, Egill and Heaton, Thomas H},
  journal   = {Bulletin of the Seismological Society of America},
  volume    = {108},
  number    = {5A},
  pages     = {2894--2901},
  year      = {2018},
  _ublisher = {GeoScienceWorld}
}

@inproceedings{le2016data,
  title     = {Data Augmentation for Time Series Classification using Convolutional Neural Networks},
  author    = {Le Guennec, Arthur and Malinowski, Simon and Tavenard, Romain},
  booktitle = {ECML/PKDD Workshop on Advanced Analytics and Learning on Temporal Data},
  _location = {Riva del Garda, Italy},
  pages     = {alshs-01357973},
  year      = {2016},
  month     = {09},
  _ublisher = {HAL},
  _address  = {Lyon, France}
}

@inproceedings{wen2022transformers,
  title     = {Transformers in Time Series: A Survey},
  author    = {Wen, Qingsong and Zhou, Tian and Zhang, Chaoli and Chen, Weiqi and Ma, Ziqing and Yan, Junchi and Sun, Liang},
  booktitle = {Proceedings of the Thirty-Second International Joint Conference on Artificial Intelligence ({IJCAI-23})},
  _location = {Macao, China},
  pages     = {6778--6786},
  year      = {2023},
  month     = {8},
  _ublisher = {IJCAI},
  doi       = {10.24963/ijcai.2023/759}
}

@article{raissi2019physics,
  title     = {Physics-Informed Neural Networks: A Deep Learning Framework for Solving Forward and Inverse Problems Involving Nonlinear Partial Differential Equations},
  author    = {Raissi, Maziar and Perdikaris, Paris and Karniadakis, George E},
  journal   = {Journal of Computational Physics},
  volume    = {378},
  pages     = {686--707},
  year      = {2019},
  month     = {2},
  _ublisher = {Elsevier},
  _address  = {Amsterdam, Netherlands},
  _issn     = {0021-9991},
  doi       = {10.1016/j.jcp.2018.10.045}
}

@article{noh2021deep,
  title         = {Deep-Learning Inversion Method for the Interpretation of Noisy Logging-While-Drilling Resistivity Measurements},
  author        = {Noh, Kyubo and Pardo, David and Torres-Verdin, Carlos},
  journal       = {arXiv preprint arXiv:2111.07490},
  year          = {2021},
  archiveprefix = {arXiv},
  primaryclass  = {physics.geo-ph},
  doi           = {10.48550/arXiv.2111.07490}
}

@article{brazell2019machine,
  title     = {A Machine-Learning-Based Approach to Assistive Well-Log Correlation},
  author    = {Brazell, Seth and Bayeh, Alex and Ashby, Michael and Burton, Darrin},
  journal   = {Petrophysics},
  volume    = {60},
  number    = {04},
  pages     = {469--479},
  year      = {2019},
  month     = {08},
  _ublisher = {SPWLA},
  _address  = {Houston, TX, USA},
  _issn     = {1529-9074},
  doi       = {10.30632/PJV60N4-2019a1}
}

@inproceedings{elhadidy2025optimizing,
  title     = {Optimizing Well Perforation with Machine Learning: A Breakthrough in Predictive Modeling},
  author    = {Elhadidy, Ahmed and Helmy, Ahmed and Heikal, Mohamed and Hany, Wael},
  booktitle = {SPE Gas \& Oil Technology Showcase and Conference},
  _location = {Dubai, United Arab Emirates},
  pages     = {D022S002R002},
  year      = {2025},
  month     = {04},
  _ublisher = {SPE},
  _address  = {Richardson, TX, USA},
  doi       = {10.2118/224556-MS}
}

@article{viggen2025improving,
  title     = {Improving Pipe Perforation Estimates from Ultrasonic Imaging Using Subpixel Machine Learning Trained on Optical Data},
  author    = {Viggen, Erlend Magnus and Grønsberg, Sondre and Brekke, Svein and Hicks, Brad and Wifstad, Sigurd Vangen},
  journal   = {Geoenergy Science and Engineering},
  volume    = {246},
  pages     = {213541},
  year      = {2025},
  _ublisher = {Elsevier},
  _address  = {Amsterdam, The Netherlands},
  _issn     = {2949-8910},
  doi       = {10.1016/j.geoen.2024.213541}
}

@article{murugan2017regularization,
  title         = {Regularization and Optimization Strategies in Deep Convolutional Neural Network},
  author        = {Murugan, Pushparaja and Durairaj, Shanmugasundaram},
  journal       = {arXiv preprint arXiv:1712.04711},
  year          = {2017},
  archiveprefix = {arXiv},
  primaryclass  = {cs.CV},
  doi           = {10.48550/arXiv.1712.04711}
}

@inproceedings{ioffe2015batch,
  title     = {Batch Normalization: Accelerating Deep Network Training by Reducing Internal Covariate Shift},
  author    = {Ioffe, Sergey and Szegedy, Christian},
  booktitle = {32nd International Conference on Machine Learning (ICML)},
  _location = {Lille, France},
  pages     = {448--456},
  year      = {2015},
  month     = {07},
  _ublisher = {PMLR},
  _address  = {Cambridge, MA, USA},
  _url      = {http://proceedings.mlr.press/v37/ioffe15.html}
}

@inproceedings{santurkar2018does,
  title     = {How Does Batch Normalization Help Optimization?},
  author    = {Santurkar, Shibani and Tsipras, Dimitris and Ilyas, Andrew and Madry, Aleksander},
  booktitle = {Neural Information Processing Systems (NeurIPS)},
  _location = {Montreal, QC, Canada},
  pages     = {2488--2498},
  volume    = {31},
  year      = {2018},
  month     = {12},
  _ublisher = {Curran Associates, Inc.},
  _address  = {Red Hook, NY, USA},
  _url      = {https://papers.nips.cc/paper_files/paper/2018/file/905056c1ac1dad141560467e0a99e1cf-Paper.pdf}
}

@inproceedings{bjorck2018understanding,
  title     = {Understanding Batch Normalization},
  author    = {Bjorck, Nils and Gomes, Carla P and Selman, Bart and Weinberger, Kilian Q},
  booktitle = {Neural Information Processing Systems (NeurIPS)},
  _location = {Montreal, QC, Canada},
  pages     = {7705--7716},
  volume    = {31},
  year      = {2018},
  month     = {12},
  _ublisher = {Curran Associates, Inc.},
  _address  = {Red Hook, NY, USA},
  _url      = {https://papers.nips.cc/paper_files/paper/2018/file/36072923bfc3cf47745d704feb489480-Paper.pdf}
}

@inproceedings{asif2020effect,
  title     = {Effect of Data Normalization on Neural Networks for the Forward Modelling of Transient Electromagnetic Data},
  author    = {Asif, Muhammad Rizwan and Bording, Thue Sylvester and Barfod, Adrian S and Auken, Esben and Larsen, Jakob Juul},
  booktitle = {NSG2020 26th European Meeting of Environmental and Engineering Geophysics},
  volume    = {2020},
  number    = {1},
  pages     = {1--5},
  year      = {2020},
  month     = {12},
  _ublisher = {EAGE},
  _address  = {Amsterdam, The Netherlands},
  _issn     = {2214-4609},
  doi       = {10.3997/2214-4609.202020061}
}

@inproceedings{yang2021delving,
  title     = {Delving into Deep Imbalanced Regression},
  author    = {Yang, Yuzhe and Zha, Kaiwen and Chen, Yingcong and Wang, Hao and Katabi, Dina},
  booktitle = {38th International Conference on Machine Learning (ICML)},
  _location = {Online},
  pages     = {11842--11851},
  year      = {2021},
  volume    = {139},
  month     = {07},
  _ublisher = {PMLR},
  _address  = {Cambridge, MA, USA},
  _url      = {http://proceedings.mlr.press/v139/yang21m/yang21m.pdf}
}

@inproceedings{szegedy2016rethinking,
  title     = {Rethinking the Inception Architecture for Computer Vision},
  author    = {Szegedy, Christian and Vanhoucke, Vincent and Ioffe, Sergey and Shlens, Jon and Wojna, Zbigniew},
  booktitle = {IEEE Conference on Computer Vision and Pattern Recognition (CVPR)},
  _location = {Las Vegas, NV, USA},
  pages     = {2818--2826},
  year      = {2016},
  month     = {06},
  _ublisher = {IEEE},
  _address  = {Piscataway, NJ, USA},
  doi       = {10.1109/CVPR.2016.308}
}

@inproceedings{stoller2018wave,
  title     = {Wave-U-Net: A Multi-Scale Neural Network for End-to-End Audio Source Separation},
  author    = {Stoller, Daniel and Ewert, Sebastian and Dixon, Simon},
  booktitle = {19th International Society for Music Information Retrieval Conference (ISMIR)},
  _location = {Paris, France},
  pages     = {334--340},
  year      = {2018},
  month     = {09},
  _url      = {https://ismir2018.ircam.fr/doc/pdfs/205_Paper.pdf}
}

@article{krizhevsky2017imagenet,
  title     = {ImageNet Classification with Deep Convolutional Neural Networks},
  author    = {Krizhevsky, Alex and Sutskever, Ilya and Hinton, Geoffrey E.},
  journal   = {Communications of the ACM},
  volume    = {60},
  number    = {6},
  pages     = {84--90},
  year      = {2017},
  month     = {05},
  _ublisher = {ACM},
  _address  = {New York, NY, USA},
  _issn     = {0001-0782},
  doi       = {10.1145/3065386}
}

@article{krizhevsky2014one,
  title         = {One Weird Trick for Parallelizing Convolutional Neural Networks},
  author        = {Krizhevsky, Alex},
  journal       = {arXiv preprint arXiv:1404.5997},
  year          = {2014},
  archiveprefix = {arXiv},
  primaryclass  = {cs.NE},
  doi           = {10.48550/arXiv.1404.5997}
}

@article{geng2016label,
  title     = {Label Distribution Learning},
  author    = {Geng, Xin},
  journal   = {IEEE Transactions on Knowledge and Data Engineering},
  volume    = {28},
  number    = {7},
  pages     = {1734--1748},
  year      = {2016},
  month     = {07},
  _ublisher = {IEEE},
  _address  = {Piscataway, NJ, USA},
  _issn     = {1041-4347},
  doi       = {10.1109/TKDE.2016.2545658}
}

@incollection{lecun2002efficient,
  title     = {Efficient Backprop},
  author    = {LeCun, Yann and Bottou, L{\'e}on and Orr, Genevieve B and M{\"u}ller, Klaus-Robert},
  booktitle = {Neural Networks: Tricks of the Trade},
  pages     = {9--50},
  year      = {2002},
  publisher = {Springer},
  address   = {Berlin, Germany},
  isbn      = {978-3-642-35289-8},
  doi       = {10.1007/978-3-642-35289-8}
}

\vfill

\section*{Biographies}

\begin{IEEEbiography}[{\includegraphics[width=1in,height=1.25in,clip,keepaspectratio]{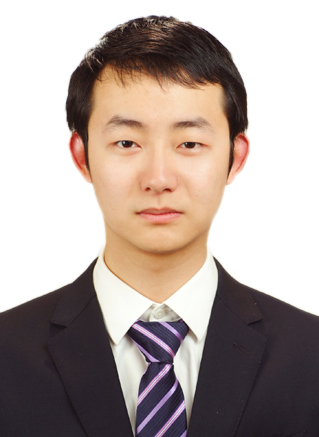}}]
~~
Si-Yu Xiao received the B.S. degree~in Microelectronic Science and Engineering from University of Electronic Science and Technology of China (UESTC) in 2018. He is currently pursuing the Ph.D. degree in Electronic Science and Technology. His research interests include digital circuit design, edge computing, artificial intelligence and their applications.
\end{IEEEbiography}

\begin{IEEEbiography}[{\includegraphics[width=1in,height=1.25in,clip,keepaspectratio]{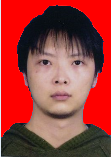}}]
~
Xin-Di Zhao received the Bachelor of Engineering degree in Software Engineering from Southwest Petroleum University. He is now a Level 2 Engineer in perforation technology research at the Southwest Branch of China National Petroleum Corporation Logging Co., Ltd. His current research includes new perforation tools, technologies and methods.
\end{IEEEbiography}

\begin{IEEEbiography}[{\includegraphics[width=1in,height=1.25in,clip,keepaspectratio]{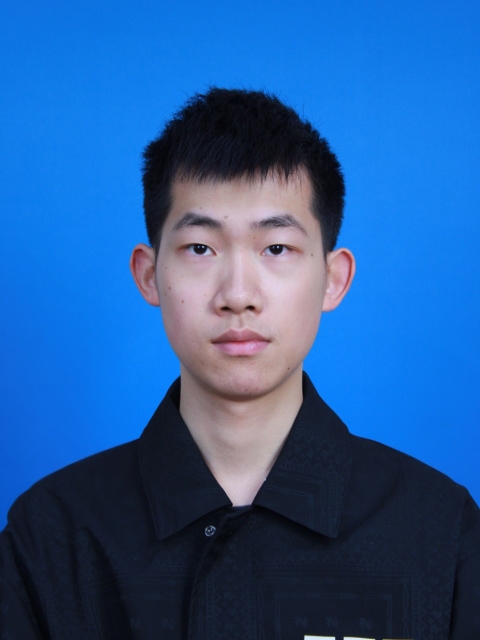}}]
~
Tian-Hao Mao is currently pursuing a Bachelor’s degree in Integrated Circuit Design and Integrated Systems at the University of Electronic Science and Technology of China. His academic interests focus on digital circuit design, system-on-chip (SoC) architecture, and algorithm optimization for hardware acceleration.
\end{IEEEbiography}

\begin{IEEEbiography}[{\includegraphics[width=1in,height=1.25in,clip,keepaspectratio]{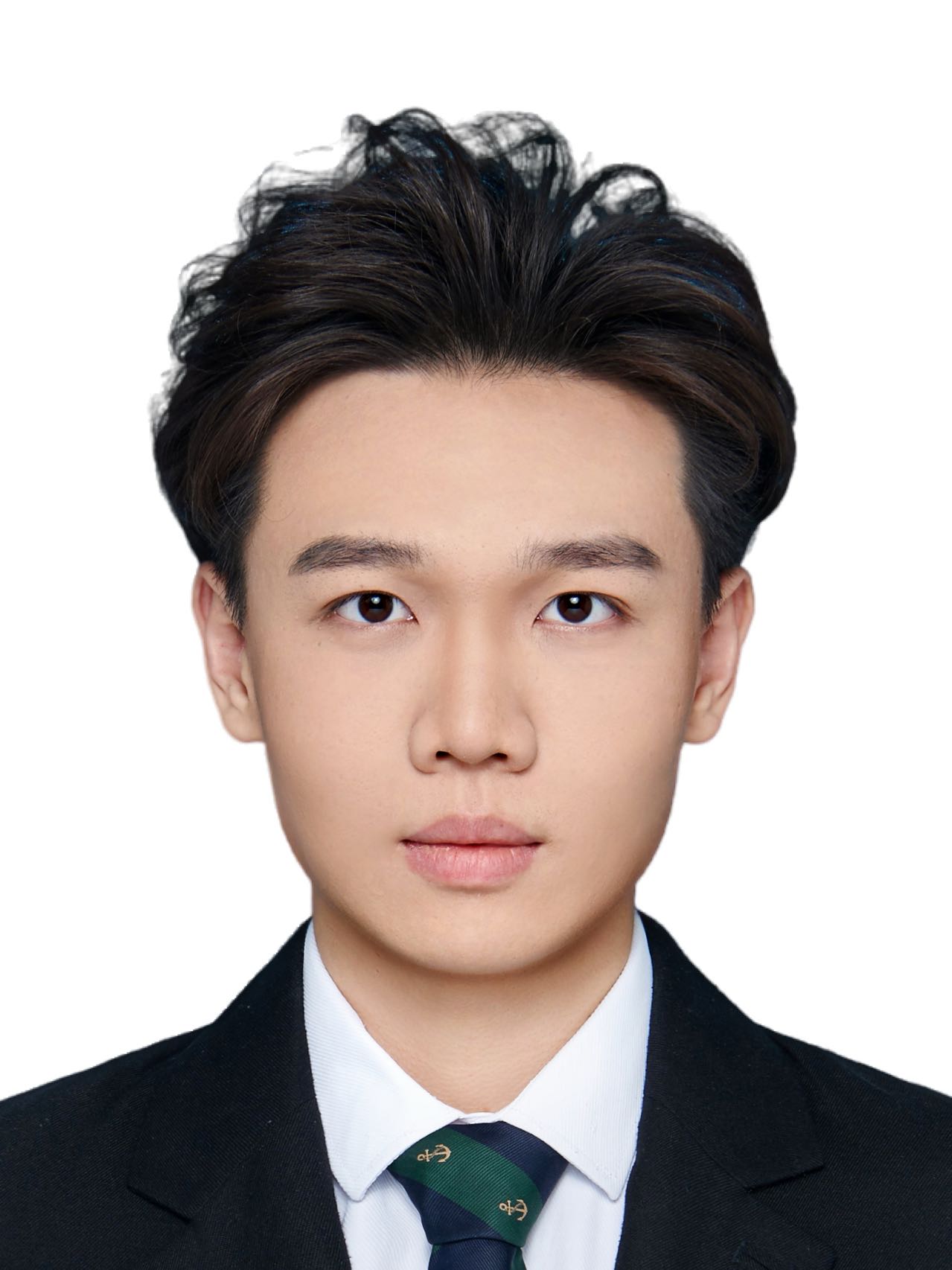}}]
~
Yi-Wei Wang received the B.S. degree in microelectronics from the University of Electronic Science and Technology of China, Chengdu, China. He is currently working toward the M.S. degree with the School of Integrated Circuits, University of Electronic Science and Technology of China. His research interests include artificial intelligence algorithms, neural network accelerator design, and their applications in artificial intelligence.
\end{IEEEbiography}

\begin{IEEEbiography}[{\includegraphics[width=1in,height=1.25in,clip,keepaspectratio]{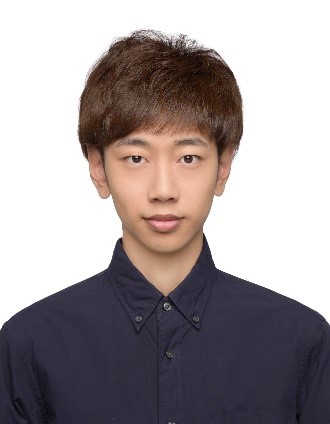}}]
~
Yu-Qiao Chen is currently pursuing the M.S. degree in Integrated Circuit Science and Engineering at the University of Electronic Science and Technology of China, where he also received the B.S. degree in Microelectronic Science and Engineering. His research interests cover digital circuit design, System-on-Chip (SoC), and neural network algorithms.
\end{IEEEbiography}

\begin{IEEEbiography}[{\includegraphics[width=1in,height=1.25in,clip,keepaspectratio]{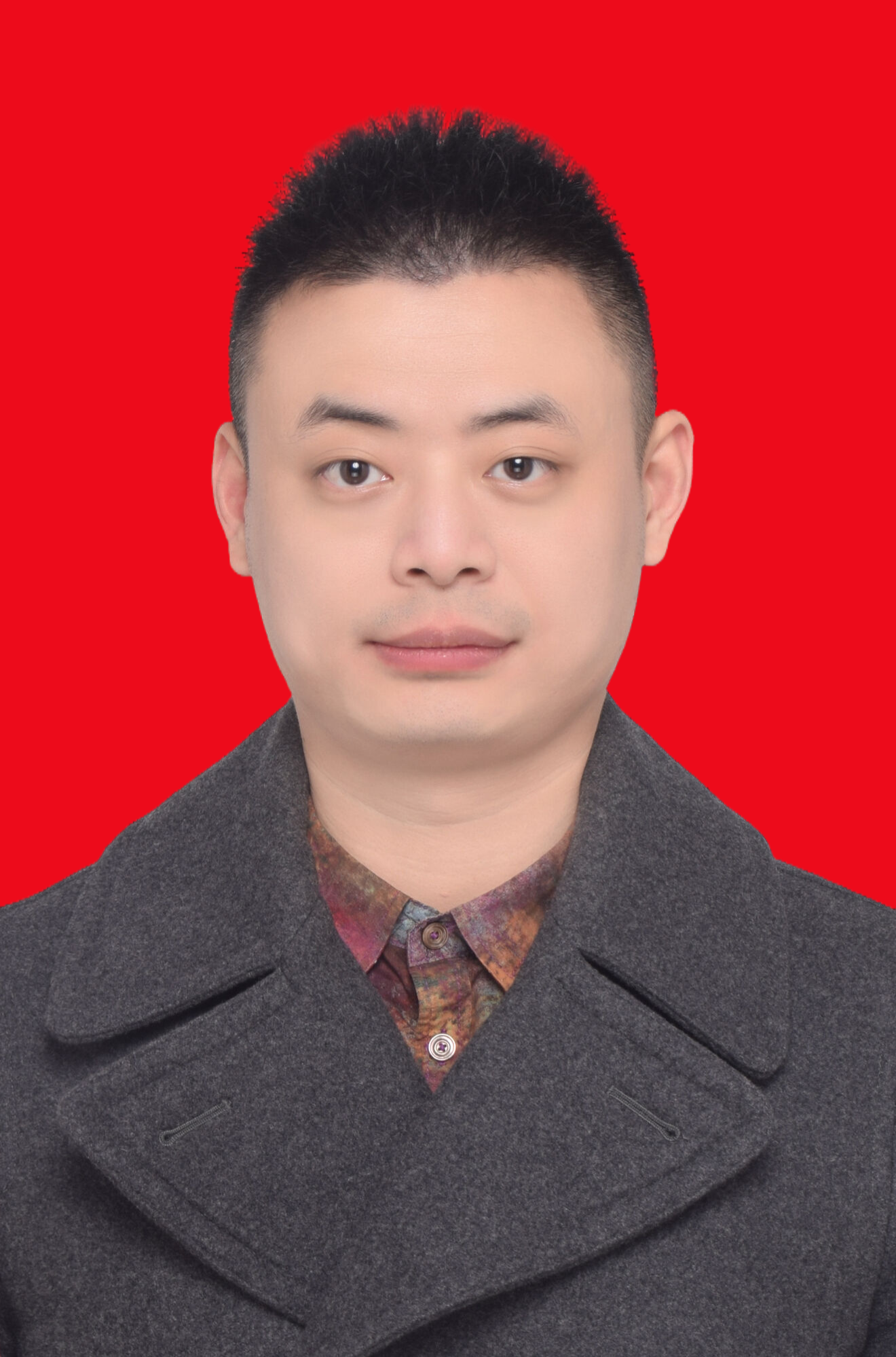}}]
~
Hong-Yun Zhang received the Master of Science degree in Communication Networks and Software from the University of Surrey, UK. He is now a Level 4 Engineer in perforation technology research at the Southwest Branch of China National Petroleum Corporation Logging Co., Ltd. His current research includes new perforation tools, technologies and methods.
\end{IEEEbiography}

\begin{IEEEbiography}[{\includegraphics[width=1in,height=1.25in,clip,keepaspectratio]{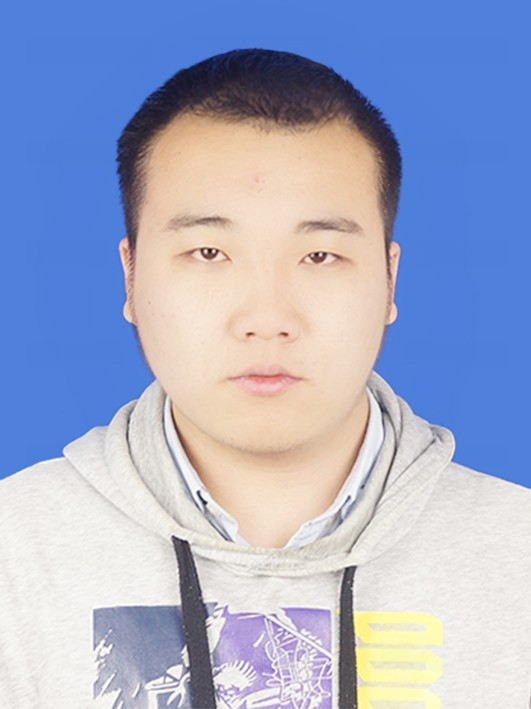}}]
~
Jian Wang received the Bachelor of Engineering degree in Software Engineering from Chengdu University of Technology. He is now a Level 4 Engineer in perforation technology research at the Southwest Branch of China National Petroleum Corporation Logging Co., Ltd. His current research includes new perforation tools, technologies and methods.
\end{IEEEbiography}

\begin{IEEEbiography}[{\includegraphics[width=1in,height=1.25in,clip,keepaspectratio]{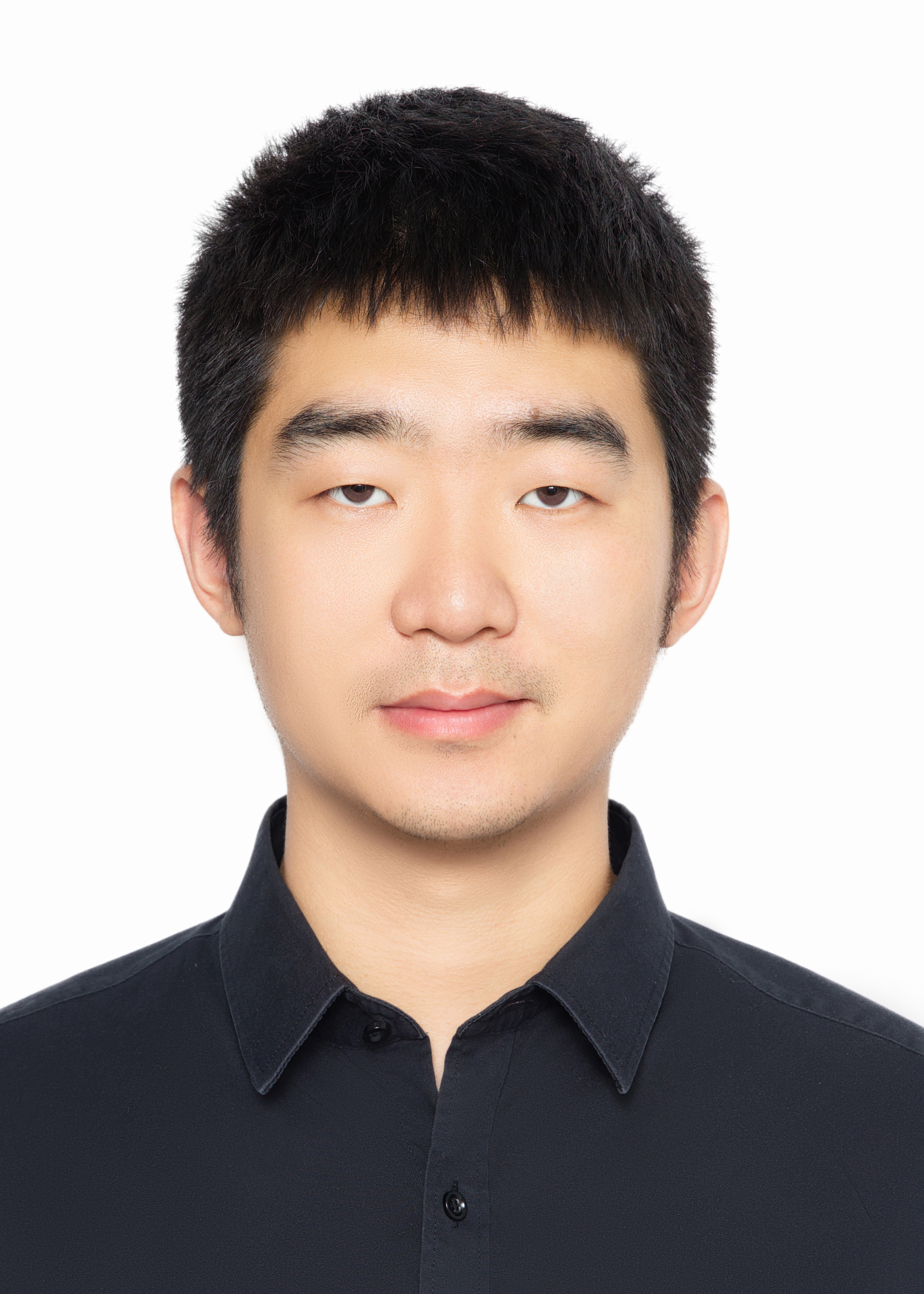}}]
~
Jun-Jie Wang received the Ph.D. degree in microelectronics from the University of Electronic Science and Technology of China. He is now a researcher at the University of Electronic Science and Technology of China. His current research interests include digital circuit design, nonvolatile memory devices, and their applications in artificial intelligence.
\end{IEEEbiography}

\begin{IEEEbiography}[{\includegraphics[width=1in,height=1.25in,clip,keepaspectratio]{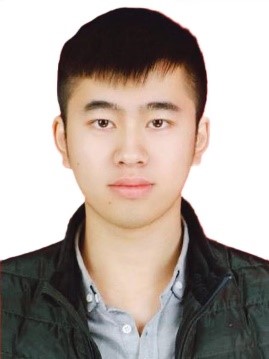}}]
~
Shuang Liu received the Ph.D. degree in microelectronics from the University of Electronic Science and Technology of China. He is now a postdoctoral researcher at the University of Electronic Science and Technology of China. His current research interests include processing{-}in{-}memory circuits and neuromorphic systems.
\end{IEEEbiography}

\begin{IEEEbiography}[{\includegraphics[width=1in,height=1.25in,clip,keepaspectratio]{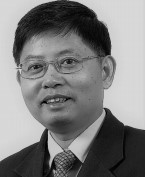}}]
~
Tu-Pei Chen is a tenured faculty member in the School of Electrical and Electronic Engineering at Nanyang Technological University, Singapore, where he has been teaching and conducting research in the field of Microelectronics for over 25 years. His current research interests include on-chip ESD and latch-up protection, memory devices, memory-based computing (in-memory computing, neuromorphic computing), and thin-film transistors and applications.
\end{IEEEbiography}

\begin{IEEEbiography}[{\includegraphics[width=1in,height=1.25in,clip,keepaspectratio]{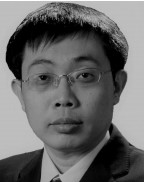}}]
~
Yang Liu received the B.Sc. degree in microelectronics from Jilin University, China, in 1998 and the Ph.D. degree from Nanyang Technological University, Singapore, in 2005. From May 2005 to July 2006, he was a Research Fellow with Nanyang Technological University, Singapore. In 2006, he was awarded the prestigious Singapore Millennium Foundation Fellowship. In 2008, he joined the School of Microelectronics, University of Electronic Science and Technology, China, as a full professor. He is the author or coauthor of over 130 peer-reviewed journal papers and more than 100 conference papers. He has been awarded one US patent and more than 30 China patents also. His current research includes memristor neural network system, neuromorphic computing ICs, and AI-RFICs.
\end{IEEEbiography}

\vfill

\end{document}